\documentclass{article} 
\usepackage{iclr2022_conference,times}

\usepackage{amsmath,amsfonts,bm}

\def\eqref#1{equation~\ref{#1}}

\def\1{\bm{1}}

\DeclareMathAlphabet{\mathsfit}{\encodingdefault}{\sfdefault}{m}{sl}
\SetMathAlphabet{\mathsfit}{bold}{\encodingdefault}{\sfdefault}{bx}{n}

\newcommand{\E}{\mathbb{E}}

\newcommand{\softmax}{\mathrm{softmax}}

\DeclareMathOperator*{\argmax}{arg\,max}

\usepackage{hyperref}
\usepackage{url}

\usepackage{booktabs}       
\usepackage{amsfonts}       
\usepackage{nicefrac}       
\usepackage{microtype}      
\usepackage{xcolor}         
\usepackage{graphicx}
\usepackage{amsmath} 

\usepackage{hyperref}
\usepackage{cleveref}

\newcommand{\algname}{Behavior Acquisition through Successor-feature Intention inference from Samples}
\newcommand{\acronym}{BASIS}
\usepackage[noend]{algpseudocode}
\usepackage{algorithm, algorithmicx}
\usepackage{etoolbox,siunitx}
\usepackage{amssymb}
\usepackage{exercise}
\usepackage{amsfonts}
\usepackage{stackengine}
\def\delequal{\stackrel{\triangle}{=}} 
\usepackage{caption}
\usepackage{subcaption}

\title{Basis for Intentions: Efficient Inverse \\ Reinforcement Learning using Past Experience}

\author{%
  Marwa Abdulhai$^1$, Natasha Jaques$^2$, Sergey Levine$^1$ \\
  $^1$Department of Computer Science, University of California, Berkeley\\
  $^2$Google Research, Brain Team \\
  \texttt{marwa\_abdulhai@berkeley.edu, natashajaques@google.com} \\ 
  \texttt{svlevine@eecs.berkeley.edu} \\
}

\iclrfinalcopy 
\begin{document}

\maketitle
\begin{abstract}
This paper addresses the problem of inverse reinforcement learning (IRL) -- inferring the reward function of an agent from observing its behavior. IRL can provide a generalizable and compact representation for apprenticeship learning, and enable accurately inferring the preferences of a human in order to assist them. 
However, effective IRL is challenging, because many reward functions can be compatible with an observed behavior. We focus on how prior reinforcement learning (RL) experience can be leveraged to make learning these preferences faster and more efficient. We propose the IRL algorithm \acronym~ (\algname), which leverages multi-task RL pre-training and successor features to allow an agent to build a strong basis for intentions that spans the space of possible goals in a given domain. When exposed to just a few expert demonstrations optimizing a novel goal, the agent uses its basis to quickly and effectively infer the reward function. Our experiments reveal that our method is highly effective at inferring and optimizing demonstrated reward functions, accurately inferring reward functions from less than 100 trajectories. 

\end{abstract}

\section{Introduction}
Inverse reinforcement learning (IRL) seeks to identify a reward function under which observed behavior of an expert is optimal. Once an agent has effectively inferred the reward function, it can then use standard (forward) RL to optimize it, and thus acquire not only useful skills by observing demonstrations, but also a reward function as an explanation for the demonstrator's behavior. By inferring the underlying goal being pursued by the demonstrator, the agent is more likely to be able to generalize to a new scenario in which it must optimize that goal, versus an agent which merely imitates the demonstrated actions. IRL has already proven useful in applications including autonomous driving, where learned models capture the behavior of nearby drivers and pedestrians \citep{huang2021driving, pedestrian_behavior}, and is a key component in enabling assistive technologies where a helper agent must infer the goals of the human it is assisting \citep{hadfieldmenell2016cooperative}.

However, IRL becomes difficult when the model does not know which aspects of the environment are potentially relevant for obtaining reward and which are distractions from achieving its intended goal. Hence, effective IRL often depends heavily on good features \citep{apprenticeship, max_entropy_irl}. Inferring relevant features from raw, high-dimensional observations is extremely challenging, because there are many possible reward functions consistent with a set of demonstrations. For this reason, previous work has often focused on IRL with hand-crafted features that manually build in prior knowledge \citep{max_entropy_irl, apprenticeship, max_margin_planning}.
When learning rewards from scratch, modern deep IRL algorithms often require a large number of demonstrations and trials \citep{garg2021iqlearn}. 

\begin{figure*}[t]
    \centering
    \includegraphics[scale=0.25]{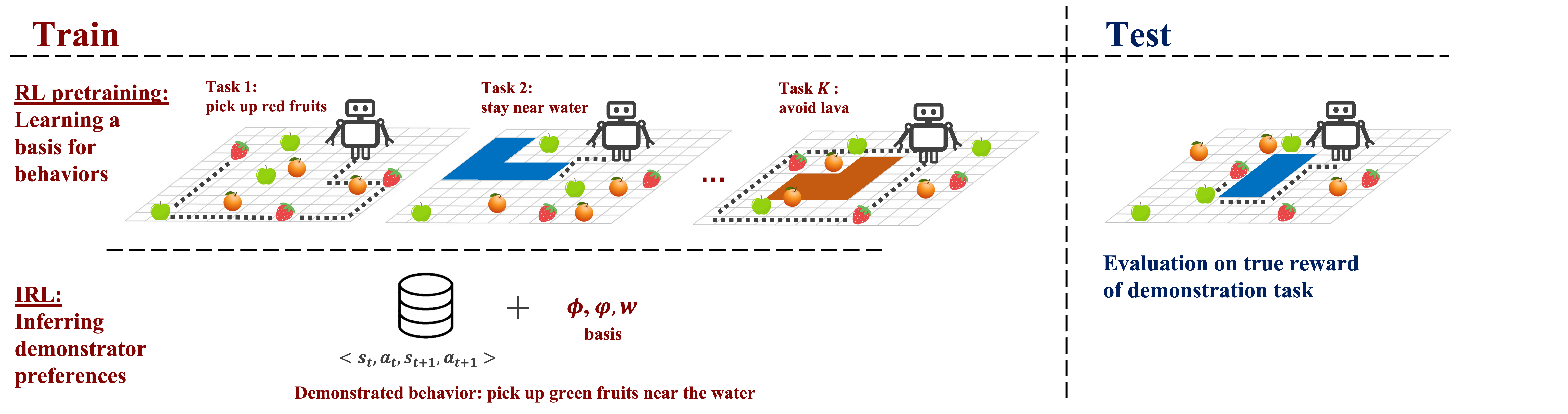}
    \centering
    \caption{\acronym{} uses 
    multi-task RL pre-training to learn a ``basis for intentions". It encodes information about both the environment dynamics, and---through modeling the rewards for multiple pre-training tasks---the space of possible goals that can be pursued in the environment. 
    It captures this information in cumulants $\phi$, successor representation $\psi$, and preference vectors $w_{1:K}$. The agent then leverages knowledge from these parameters to rapidly infer the demonstrator's goal shown through demonstrations $(s_{t}, a_{t}, s_{t+1})$, updating the parameters as needed.}
    \label{fig:approach}
\end{figure*}

In contrast, humans quickly and easily infer the intentions of other people. As shown by \citet{qian2021modeling}, humans can infer rewards more effectively than our best IRL algorithms, as they bring to bear strong prior beliefs as to what might constitute a reasonable goal -- e.g., that a person moving towards a wall is more likely to have the intention of turning off the light as opposed to moving to a random point. This skill comes from humans having access to a lot of previous experience successfully accomplishing prior goals or watching others pursue their preferences \citep{goal_inference_planning}. We hypothesize that prior knowledge of the space of probable goals is important to effectively and efficiently infer intentions with IRL. As shown by \citet{Ng00algorithmsfor, apprenticeship, max_margin_planning}, IRL methods that utilize user-supplied features concisely capture the structure of the reward function. We hypothesize the path towards building scalable IRL methods entails being able to instead \textit{learn} those features from past experience. Thus, we propose an IRL algorithm, \acronym{} (\algname), that leverages multi-task RL pre-training and successor features to enable the agent to first learn a \textit{basis for intentions} that spans the space of potential tasks in a domain.  Using this basis, the agent can then perform more efficient IRL or inference of goals. Figure \ref{fig:approach} shows an overview of our approach.

We use successor features to enable learning a basis for intentions because they allow learning a representation that naturally decouples the dynamics of the environment from the rewards of the agent, which are represented with a low-dimensional preference vector \citep{dayan_successor, barreto2018successor, filos2021psiphilearning}. Via multi-task pre-training, the agent learns a representation in which the same successor features are shared across multiple tasks, as in \citet{barreto2018successor}. When the agent is tasked with inferring the rewards of a novel demonstrator via IRL, it initializes its model of the other agent with the learned successor features and a randomly initialized preference vector. Thus, the agent starts with a strong prior over the environment dynamics and the space of reasonable policies. It can then quickly infer the low-dimensional preference vector, while updating the successor features, in order to accurately describe the demonstrated behaviour.

We evaluate \acronym~in three multi-task environment domains: a gridworld environment, a highway driving scenario, and a roundabout driving scenario. On these tasks, our approach is up to 10x more accurate at recovering the demonstrator's reward than state-of-the-art IRL methods involving pre-training with \textit{IRL}, and achieves up to 15x more ground-truth reward than state-of-the-art imitation learning methods.
In summary, the contributions of this paper are to show the effectiveness of multi-task RL pre-training for IRL, to propose a new technique for using successor features to learn a basis for behaviour that can be used to infer rewards, and empirical results demonstrating the effectiveness of our method over prior work.

\section{Related work}

\textbf{Inverse RL:} 
IRL methods learn the reward function of an agent through observing expert demonstrations. Depending on whether the goal is imitation, explanation, or transfer, downstream applications might use the recovered policy, reward function, or both.
In environments with high-dimensional state spaces, there are many possible reward functions that are consistent with a set of demonstrations. Thus, early work on IRL relied on hand-engineered features that were known to be relevant to the reward function \citep{Ng00algorithmsfor,apprenticeship,max_margin_planning,syed2008apprenticeship, NIPS2010_a8f15eda}. 
\citet{max_entropy_irl} propose maximum entropy IRL, which assumes that the probability of an action being seen in the demonstrations increases exponentially with its reward, an approach we also follow in this paper. 
A series of deep IRL methods have emerged that learn reward structures \citep{jin2017inverse, burchfiel_scoring_traj, shiarlis_irl_failure} or features \citep{wulfmeier2016maximum, 8593438, JARAETTINGER2019105,fu2018learning} from input. IQL \citep{kalweit2020deep} is a recent IRL method which uses inverse-action value iteration to recover the underlying reward. It was benchmarked against \citep{wulfmeier2016maximum} and found to provide superior performance, so we compare to IQL in this paper. 
However, these prior methods do not leverage past multi-task RL experience as a way to overcome the underspecification issue to improve IRL. 

Meta-inverse reinforcement learning methods, including those proposed by \citet{yu2019metainverse, xu2019learning, gleave2018multitask}, have applied meta-learning techniques to IRL, by pre-training on past \textit{IRL} problems, then using meta-learning to adapt to a new IRL problem at test time while leveraging this past experience. 
In contrast with meta-IRL methods, we pre-train using \textit{RL}, in which the agent has a chance to explore the environment and learn to obtain high rewards on multiple tasks. Relying on RL rather than IRL pre-training provides an advantage, since collecting the demonstrations required for IRL can be expensive, but RL only requires access to a simulator. Further, the basis learned through RL enables our agent to rapidly infer rewards in demonstrated trajectories, even in complex and high-dimensional environments.

\textbf{Imitation learning:} 
IL methods attempt to replicate the policy that produced a set of demonstrations. The most straightforward IL method is behavioural cloning \citep{abs-1011-0686, Bain96aframework}, where the agent receives training data of encountered states and the resulting actions of the expert demonstrator, and uses supervised learning to imitate this data \citep{pmlr-v9-ross10a}. This allows the agent observing the expert data to learn new behaviors without having to interact with the environment. \citet{Kostrikov2020Imitation, NEURIPS2020_524f141e, chan2021scalable} have proposed several other IL methods.
Some IL methods allow the agent to interact with the environment in addition to receiving demonstrations, and include adversarial methods such as \citet{NIPS2016_cc7e2b87, kostrikov2018discriminatoractorcritic, baram2016modelbased}. Similar to \citet{borsa2017observational, torabi2018behavioral, sermanet2018timecontrastive, liu2018imitation, brown2019extrapolating}, our approach leverages learning from demonstrations where actions are available but rewards and task annotations are unknown. One among these methods is the non-adversarial method IQ-Learn \citep{garg2021iqlearn}, which we compare with our method.
However, current IL methods that focus on skill transfer have a different goal of minimizing the supervised learning loss and do not recover the reward function, which \acronym{} does.

\textbf{Successor features:} 
\citet{barreto2018successor} derives generalized successor features from successor representations \citep{dayan_successor} to leverage reward-free demonstrations
for learning cumulants, successor features, and corresponding preferences. These have been used for applications including planning \citep{zhu2017visual}, zero-shot transfer \citep{lehnert2017advantages, barreto2018successor, borsa218, Barreto2020FastRL, filos2021psiphilearning}, exploration \citep{NEURIPS2019_1b113258, Machado_Bellemare_Bowling_2020}, skill discovery \citep{machado2018eigenoption, hansen2020fast}, apprenticeship learning \citep{batch_apprenticeship_successor}, and theory of mind \citep{pmlr-v80-rabinowitz18a}. 
PsiPhi-Learning \citep{filos2021psiphilearning} illustrates that generalized value functions, such as successor features, are a very effective way to transfer knowledge about agency in multi-agent settings, and includes an experiment which uses successor features for IRL. 
Unlike PsiPhi, which learns a new set of successor features for each agent it models, we learn a shared set of successor features spanning all tasks that have been seen in the multi-task learning phase, similar to \citet{barreto2018successor}; this enables more effective transfer to new tasks that have not been encountered during training. 
However, \citet{barreto2018successor} does not address IRL or learning from demonstrations. 
Through this protocol, we demonstrate a notion of basis for intentions learned from past experience that can be leveraged to quickly infer intentions.

\section{Background and problem setting}

\textbf{Markov decision processes:}
An agent's interaction in the environment can be represented by a Markov decision process (MDP) ~\citep{puterman1994markov}. 
Specifically, an MDP is defined as a tuple $\mathcal{M}\!=\!\langle \mathcal{S}, \mathcal{A}, P, \mathcal{R}, \gamma \rangle$;
$\mathcal{S}$ is the state space, 
$\mathcal{A}$ is the action space, 
$P\!:\!\mathcal{S}\times\mathcal{A}\mapsto \mathcal{S}$ is the state transition probability function,
$\mathcal{R}$ is the reward function, and
$\gamma\!\in\![0,1)$ is the discount factor.
An agent executes an action at each timestep $t$ according to its stochastic policy $a_t\!\sim\!\pi(a_{t}|s_{t})$, where $s_{t}\!\in\!\mathcal{S}$.
An action $a_t$ yields a transition from the current state $s_{t}$ to the next state $s_{t+1}\!\in\!\mathcal{S}$ with probability $P(s_{t+1}|s_t,a_t)$. 
An agent then obtains a reward according to the reward function $r_t\!\sim\!\mathcal{R}(s_t,a_t)$.
An agent's goal is to maximize its expected discounted return $\mathbb{E}\left[\sum_t \gamma^t r_t|s_{0},a_0\right] = Q(s_0,a_0)$. 

\textbf{Successor features:}
decouple modeling the dynamics of the environment from modeling rewards \cite{barreto2018successor, dayan_successor}. The value function is decomposed into features that represent the environment transition dynamics, and preference vectors (which are specific to a particular reward function). Thus, SFs enable quick adaptation to optimizing a new reward function in the same environment (with the same transition dynamics).
Specifically, we can represent the one-step expected reward 
as: 
\begin{equation}
\label{eq:phireward}
r(s_t, a_t, s_{t+1}) = \phi(s_t, a_t, s_{t+1})^{\top}w, 
\end{equation}
where $\phi(s_t, a_t, s_{t+1}) \in \mathcal{R}^d$ are features (called cumulants) of $(s_t,a_t, s_{t+1})$ and $w \in \mathcal{R}^d$ are weights or preferences. The preferences $w$ are a representation of a possible goal with a particular reward function, with each $w$ giving rise to a separate task. The terms ‘task’, ‘goal’, and ‘preferences’ are used interchangeably when context makes it clear whether we are referring to $w$ itself. The state-action value function for a particular policy $\pi$
can now be decomposed with the following linear form \citep{barreto2018successor}:
\begin{equation}
\begin{aligned}
    \label{eq:sf}
    Q^{\pi}(s, a) & = \psi^{\pi}(s,a)^{\top}w, \\
    \psi^{\pi}(s,a) & = \E_{[s_t = s, a_t = a]} \sum_{i=t}^{\infty} \gamma^{i-t}\phi(s_{i+1}, a_{i+1}, s_{i+1}) \\
\end{aligned}
\end{equation}
where $\gamma \in [0, 1)$, 
and $\psi^{\pi}(s,a)$ are the successor features of $\pi$. 

\textbf{Maximum-entropy IRL:}
Given a set of demonstrations $D = {(s_0, a_0), (s_1, a_1) \dots}$ provided by an expert, the IRL problem \citep{Ng00algorithmsfor} is to uncover the expert's unknown reward function $\mathcal{R}$ that resulted in the expert's policy, which in turn led to the provided demonstrations.
\citet{max_entropy_irl} propose the maximum entropy IRL framework, under which highly rewarding actions are considered exponentially more probable in the demonstrations, an assumption which we follow in this work. MaxEnt IRL states the expert's preference for any given trajectory between specified start and
goal states is proportional to the exponential of the reward along the path: $P(s,a |r)\!\propto\!\exp \{\sum_{s,a} \mathcal{R}_{s,a}\}$. Thus, the maximum entropy model proposed by \citet{max_entropy_irl} is:
\begin{equation}
\begin{aligned}
    \label{eq:max_entropy_transformation}
    P(a|s_i) = \text{exp}(Q(s_i,a))
\end{aligned}
\end{equation}
The optimal parameters can be found by maximizing the log likelihood with respect to the parameters of the reward, often utilizing either a tabular method similar to value iteration, or approximate gradient estimators based on adversarially trained discriminators.

\section{\acronym: \algname}
We now present our IRL algorithm, \acronym{}, 
which is illustrated in Figure \ref{fig:approach}. First, the agent learns a basis for intentions using successor features and multi-task RL pre-training. Then, the agent uses the successor representation learned from pre-training as an initialization for inferring the reward function of an expert from demonstrations with IRL. The successor features learned via pre-training act like a prior over intentions, enabling the agent to \textit{learn} the features for linear MaxEnt IRL so that we get the simplicity and efficiency benefits of good features, without having to specify those features manually. These successor features can then be refined to infer a likely intention using IRL based on the provided demonstrations and additional online experience.

\subsection{RL pre-training: Learning a basis for intentions}
In order to form a good basis for intentions, we use multi-task RL pre-training and successor features to learn a representation that enables the agent to solve a large variety of tasks. We assume that each of the tasks share the same state space and transition dynamics, but differ in their reward functions.
These assumptions are relevant to the setting in which a human demonstrator may have one of several possible goals in the same environment. 

To learn a representation of the tasks, we learn a global cumulant function $\phi$, successor features $\psi$, and per-task preference vectors $w_{1:K}$. As the agent operates under the same state space and transition dynamics across the different tasks, we can share $\phi$ across tasks, which enables learning a common basis for reward functions. We can think of $\phi$ as a task-agnostic set of state features that are relevant to predicting rewards across any training task. The agent's policy is captured by $\psi$, which estimates the future accumulation of these state features according to \cref{eq:sf}. We learn a separate preference vector $w_{1:K}$ for each of the $K$ tasks, which enables representing the different reward functions of the tasks. For simplicity, we refer to a preference vector specific for a task as $w$.


\begin{figure}[t]
    \centering
    \includegraphics[scale=0.80]{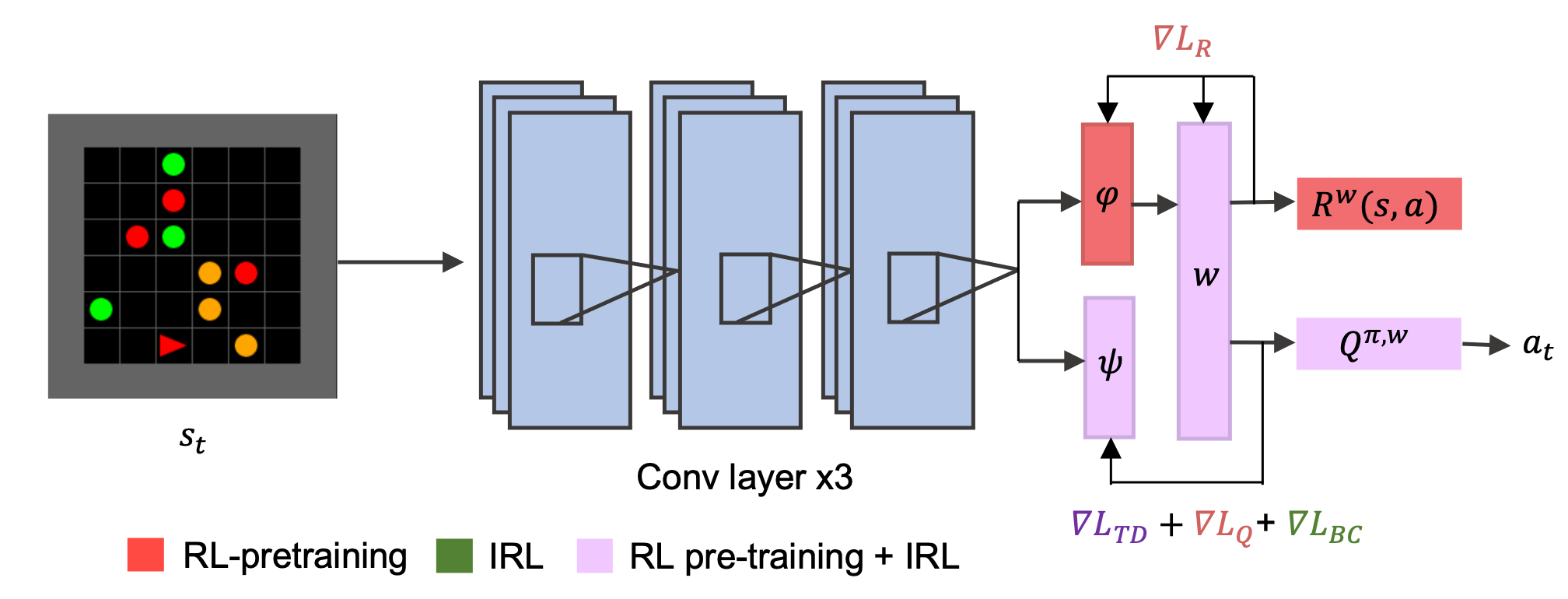}
    \centering
    \caption{Architecture diagram for learning global cumulants $\phi$ and successor features $\psi$, both of which we represent as function approximators, and task-specific preference vectors $w$. The input $s_t$ is passed through shared convolution layers highlighted in blue, which extract information about the state from pixels. This intermediate representation is passed into networks for $\psi$ and $\phi$ respectively. Taking the dot product of the output from $\psi$ and $w$, we obtain the action-value function $Q$. The dot product of $\phi$ and $w$ gives us the predicted reward at $s_t$. Networks that are updated in both RL pre-training and during IRL are highlighted in purple ($\psi, w$) and those learned solely with RL are highlighted in red. Similar color conventions are used to represent the loss functions used for each network parameter, with losses only computed during IRL highlighted in green.}
    \label{fig:networks}
\end{figure}

We use a neural network to learn both $\psi$ and $\phi$, as shown in Figure \ref{fig:networks}.
Initial features are extracted from raw, high-dimensional observations $s_t$ via a shared trunk of convolution layers. We then use separate heads output both $\phi$ and $\psi$, which are parameterized by $\theta_{\phi}$ and ${\theta_{\phi}}$, respectively. The preference vectors $w_{1:K}$ do not depend on the state, and are learned separately.
As shown in Eq. \ref{eq:sf}, combining $\psi$ with a particular preference vector $w$ produces $Q^{\pi, w} = \psi^{\pi}(s,a)^{\top}w$, which can be used as a policy $\pi$. In order to ensure our representation is suitable for later IRL training, we fit the $Q$ function using a modified version of the Bellman error with a softmax function, following the formulation of MaxEnt IRL given in \citet{max_entropy_irl}: 
\begin{equation}
\begin{split}
    \label{eq:q}
    \mathcal{L}_{\text{Q}}(\theta_{\psi}) \delequal \E_{(s_t, a_t, s_{t+1}, r_t) \sim \mathcal{B}}[ ||Q^{\pi, w}(a_t, s_t;  \theta_{\psi}) -  r_t- \gamma \,\underset{a_{t+1}}{\softmax} \space Q^{\pi, w}(s_{t+1}, a_{t+1}; \Tilde{\theta_{\psi}})||],
\end{split}
\end{equation}
where $\mathcal{B}$ represents the replay buffer. Note that this formulation of Q-learning is equivalent to Soft Q-Learning \cite{haarnoja2017reinforcement}, which is a maximum entropy RL method that can improve exploration, and is thus a reasonable choice for a forward RL objective.
 
To ensure that environment features extracted by $\phi$ are sufficient to represent the space of possible reward functions, and that each $w$ accurately represents its task-specific reward function, we train both using the following reward loss:
\begin{equation}
\begin{split}
    \label{eq:reward}
    \mathcal{L}_{\text{R}}(\theta_{\phi}, w) \delequal \E_{(s_t, a_t, r_t) \sim \mathcal{D}}||\phi(s_t, a_t; \theta_{\phi})^{\top} w\!-\!r_t||.
\end{split}
\end{equation}
As shown in Eq. \cref{eq:sf}, the successor features $\psi$ should represent the accumulation of the cumulants $\phi$ over time. To enforce this consistency, we train 
$\theta_{\psi}$ with the following inverse temporal difference (ITD) loss~\citep{barreto2018successor}, which is similar to a Bellman consistency loss:
\begin{equation}
\begin{split}
    \label{eq:itd}
    \mathcal{L}_{\text{TD}}(\theta_\psi) \delequal \E_{(s_t, a_t, s_{t+1}, a_{t+1}) \sim \mathcal{B}}[|| \psi(s_t, a_t; \theta_{\psi}) - \phi(s_t, a_t; \theta_{\phi}) - \gamma \psi(s_{t+1}, a_{t+1}; \Tilde{\theta}_{\psi})||].
\end{split}
\end{equation}
We do not train $\theta_{\phi}$ with this loss (the gradient is not passed through $\phi(s_t, a_t; \theta_{\phi})$). This is because we first force $\phi$ to represent the rewards through Eq. $\ref{eq:reward}$, then construct $\psi$ out of the fixed $\phi$, which leads to more stable training.
Through this process of successor feature learning, our method learns a ``basis for intentions" that can be used as an effective prior for IRL in the next phase. 

\subsection{Inferring intentions with IRL}
\label{sec:phaseii}
Given an expert demonstrating its preferences, the goal of IRL is to determine the intentions of this demonstrator by not only recovering its policy $\pi_{e}(a|s)$ but accurately inferring its reward function. 
Our agent is given access to these demonstrations without rewards, denoted $D = \tau_1, \tau_2 \dots \tau_N$, where the trajectory $\tau \delequal (s_0, a_0, s_1 \dots, s_T , a_T, s_{T+1})$ is generated by the demonstrator. The demonstrator is optimizing an unknown, ground-truth reward function $r_e(s_t,a_t)$ that was not part of the pre-training tasks.

We will now clarify how successor features ~\citep{barreto2018successor} can be related to the demonstrator's policy and reward function by drawing a parallel between successor features and MaxEnt IRL ~\citep{max_entropy_irl}.
This motivates our formulation of IRL with successor features. 

MaxEnt IRL assumes that the demonstrator's actions are distributed in proportion to exponentiated Q-values, i.e., $\pi(a|s) \propto \exp(Q(s,a))$ (Eq. \ref{eq:max_entropy_transformation}). We then learn the parameters of the expert's Q-function, $\theta_e$, by solving the following optimization problem: $\theta_{e}^{*} = \argmax_{\theta_e} \sum_{a=1}^{a_T} \log P(a|\pi_{e})$.
Since we can represent the expert's Q-value with successor features $\psi_e$ and preference vector $w_e$ (Eq. \ref{eq:sf}), we can express
the expert's policy as: 
$\pi_{e}(a|s) \propto \exp(\psi_e^{\top}w_e)$. 
This leads to the following behavioral cloning (BC) loss to fit our task-specific preferences $w_e$ and successor features $\psi_e$:
\begin{align}
\label{eq:bc}
\mathcal{L}_{\text{BC}}(\theta_{\psi_e}, \omega_e) \delequal  - \E_{\tau \sim D} \log  \frac{\exp(\psi_e(s, a)^{\top} w_e)}{\sum_a (\exp \psi_e(s, a)^{\top}w_e).}
\end{align}

Note however that this BC loss alone is insufficient to produce an effective IRL method, since we have no way to infer the reward function of the expert. To predict the expert's rewards, we need to make use of $\phi$; i.e.:
\begin{align}
\label{eq:constraint}
\phi_e(s, a)^{\top}w_e = r_e(s,a)
\end{align}
To ensure that $\psi_e$ and $\phi_e$ are consistent, we also require an ITD loss:
\begin{equation}
\begin{split}
    \label{eq:itd_phase2}
    \mathcal{L}_{\text{TD}}({\theta_\psi}_e) \delequal \E_{\mathcal{D}}  || \psi_e(s_t, a_t; {\theta_\psi}_e) - \phi_e(s_t, a_t; {\theta_\phi}_e) - \gamma \psi_e(s_{t+1}, a_{t+1}; {\Tilde{\theta_\psi}}_e)||.
\end{split}
\end{equation}

Now we can draw a direction connection to MaxEnt IRL \citep{max_entropy_irl},
which proposes inferring the demonstrator's reward function using a linear transformation applied to a set of state features: $\mathcal{R}(f(s);\theta) = \theta^T f(s)$ where the mapping $f: \mathcal{S} \rightarrow [0, 1]^{k}$ is a state feature vector, $\theta$ are the model parameters, and $\mathcal{R}(f(s), \theta)$ is the reward function.  
Our approach instead uses successor features to learn a set of continuous state features $\phi_e$ to replace $f$, and $w_e$ is analogous to learning $\theta$. Thus, according to the MaxEnt IRL model, if $\psi_{e}^\pi(s,a)$ remains Bellman consistent with $\phi_e$ (by minimizing the ITD loss) and $w_e$ and $\psi_e^{\pi_e}(s,a)$ are optimized so as to maximize the probability of the observed demonstration actions (as in Eq. \ref{eq:bc}), we will have recovered the demonstrator's Q-function as $\psi_e(s,a)^T w$, and the demonstrator's reward function as $\phi_e(s,a)^T w_e$. 

\textbf{Benefitting from RL pre-training:}
To initialize $\psi_e$ and $\phi_e$, the agent uses the parameters $\theta_{\phi}$ and $\theta_{\psi}$ that it learned during RL pre-training; i.e. ${\theta_\psi}_e \leftarrow \psi$ and ${\theta_\phi}_e \leftarrow \theta_{\phi}$. It also initializes a new preference vector $w_e$ as the average of all $w$ vectors across tasks learned in during RL pre-training, in order to begin with an agnostic representation of the demonstrator's goals. 
To give some intuition for the method, the $\phi$ learned during RL training represents a shared feature space that can be used to explain all the pre-training tasks. 
When the agent initializes $\phi_e$ with $\phi$, it is given a strong prior that can help explain the expert's behavior. Hence, even though at the beginning of the IRL stage the agent has never been directly trained on behavior for the new task, it can potentially extrapolate from the learned basis in order to quickly ascertain the demonstrator's preferences. To the extent that the pre-trained successor features $\psi$ provide a good basis for the expert's policy, then this learning process might primarily modify $w_e$, and only make minor changes to $\psi_e$ to make it consistent with the new policy. In this case, the method can recover the reward and policy for the new task very quickly. However, the astute reader will notice that the basis for intentions learned during RL pre-training and encoded in $\theta_{\phi}$ and $\theta_{\psi}$ does not necessarily constitute a policy that is optimal for the new task that is being demonstrated by the expert. 
Hence, the ITD loss (Eq. \ref{eq:itd}) is necessary to learn the correct $\psi_e$. Even if the demonstrator policy differs from the $\psi$ learned during RL pre-training, ITD allows us to accurately infer the demonstrated policy.

\begin{algorithm}[t]
	\caption{RL pre-training: learning a basis}
	\label{alg:basis}
	\begin{algorithmic}[1]
	   \State \text{Initialize} $\theta_{\phi}$ for cumulants, $\theta_{\psi}$ for successor features, and preference vectors ${w^{k}}_{k=1}^K$ 
       \State $B = []$ \Comment{Replay buffer initialization}
      \For{each iteration}
        \State Get initial observation $s_t$
        \State Sample task $k$ from $K$ tasks
        \For{each environment step}
            \State Get action $a \sim \pi$ 
            \State Take action $a$ and observe $x'$ and $r$
            \State $B \gets B \cup \{x, a, r, x', k\}$
      \EndFor
      \For{each gradient step} 
      \State Update $\theta_{\psi}, w$ with bellman error in \cref{eq:q}
      \State Update $\theta_{\phi}, w$ with reward loss in \cref{eq:reward}
      \State Update $\theta_{\psi}, w$ with ITD loss in \cref{eq:itd}
      \EndFor
    \EndFor
	\end{algorithmic}  
\end{algorithm}

\begin{algorithm}
	\caption{IRL: Inferring Intentions}
	\label{alg:infer}
	\begin{algorithmic}[1]
	   \State \textbf{Input:} expert demonstrations $\mathcal{D}$
	   \State \text{Initialize} ${\theta_\phi}_e \leftarrow \phi$ ${\theta_\psi}_e \leftarrow \psi$, $w_{e} \leftarrow \sum_1^{K} w_k/K$ from multi-task RL pre-training 
      \For{each demonstration $<\!\!s_{t}, a_{t}, s_{t+1}, a_{t+1}\!\!>$}:
        \State Update ${\psi}_e, w_e$ with BC loss in \cref{eq:bc}
        \State Update $\psi_e, w_e$ with ITD loss in \cref{eq:itd_phase2}
      \EndFor
	\end{algorithmic}  
\end{algorithm}

\textbf{Algorithm summary:}
We now summarize the procedure for both RL pre-training (learning a basis for intentions) shown in \cref{alg:basis} and IRL (inferring intentions) shown in \cref{alg:infer}. Algorithm \ref{alg:basis},
uses multi-task RL training, and computes the Bellman error, reward loss, and ITD loss at every gradient step to discover a good basis for intentions. In Algorithm \ref{alg:infer} (IRL), we begin the learning process with parameters initialized from the RL pre-training. We iterate through a batch of a fixed set of demonstrations from $D$, computing the BC loss and ITD loss to maintain consistency between $\psi$ and $\phi$ as aforementioned.  
At test time, we use the inferred cumulants $\phi_e$, preference vector $w_e$, and successor representation $\psi_e$, to produce a policy that can be executed in the test environment, and measure how well it matches the demonstrator's reward (as in Figure \ref{fig:approach}).

\section{Experiments}

Below we describe the research questions that our empirical experiments are designed to address. 

\textbf{Question 1: Will \acronym{} acquire the behaviors of a demonstrator expert more quickly and effectively than conventional IRL and imitation learning methods?} \\
A central goal of IRL is to be able to reproduce the demonstrated behavior of the expert in a generalizable way. We hypothesize that the RL pre-training and successor representation of \acronym{} will allow it do this more accurately and with fewer demonstrations than existing techniques. To measure how well a method can acquire demonstrated behaviors,
we evaluate performance using the \textit{expected value difference}. This metric evaluates the sub-optimality of a policy trained to optimize the reward function inferred with IRL. It is computed as the difference between the return achieved by the expert policy and the policy inferred with IRL, both measured under the ground truth reward function (thus, a lower value difference is better). 
Intuitively, this metric captures how much worse the policy that IRL recovers is vs. the expert demonstrator's policy, and is the metric of choice in for evaluating IRL methods in prior work~\citep{NIPS2011_c51ce410, wulfmeier2016maximum, xu2019learning}. For each algorithm, we evaluate the performance with different numbers of demonstrations to study whether the use of prior knowledge from other tasks in \acronym{} allows it to perform IRL more efficiently (i.e., with fewer demonstrations).

\textbf{Question 2: Can \acronym~more accurately predict the true reward with fewer demonstrations?}
It is often easier to optimize for the correct behavior of a demonstrator agent than accurately predict its reward function. Even with inaccurate reward values, the agent could still demonstrate the correct behavior as long as it estimated the relative value of different actions correctly. 
Hence, we perform further analysis to observe how well the agent is able to predict the expert's true reward function. We compute the mean squared error between the agent's prediction of the reward, and the true environment reward that is observed: $MSE = (\phi_e(s_t,a_t)^T w_e - r_t)^2$.
This allows us to understand how accurately the agent is able to infer intentions by leveraging its basis for intentions.

\textbf{Question 3: How closely does \acronym~match the demonstrated policy and adhere to the demonstrator’s preferences?}
In order to understand how well the agents are able to adhere to the demonstrators preferences, we visualize the behaviors of the agents. For example, we measure the distribution of behaviors for each method vs. the distribution of the expert's policy.

\textbf{Question 4: How does multi-task pre-training \& successor features benefit IRL?} 
We address this question through ablations that show how much multi-task RL pre-training vs successor features contribute to learning an IRL task. We compare BASIS to two ablations. The first is uses no multi-task pre-training, but does perform IRL via BC and successor features. Denoted as "No pre-training (BC + successor features)", it is used to assess the importance of RL pre-training. The second, "No successor features (pre-train with DQN)", is an algorithm which performs multi-task pre-training via DQN and IRL via BC, and assesses the importance of successor features.

\subsection{Domains}
We evaluate \acronym{} on two domains: 1) a gridworld environment Fruit-Picking, which allows us to carefully analyze and visualize the performance of the method, and 2) high-dimensional autonomous driving environments Highway and Roundabout, which necessitate the use of deep IRL. We modified the domains to be able to create multiple tasks with differing reward functions, enabling us to test generalization to novel demonstrator tasks outside of the set of pre-training tasks. Further details are available in Appendix \ref{sec:appendix}.

\textbf{Fruit-Picking} is a custom environment based on \cite{gym_minigrid}, with different colored fruits the agent must gather. In each task, the number and type of fruits varies, along with the reward received for gathering a specific type of fruit. During RL pre-training, the agent learns to pick one type of fruit per task. For the IRL phase, the demonstrated task is different to the training tasks; namely, the agent shows a varying degree of preference to each of the fruits in the environment i.e. 80\% preference for red fruits, 20\% preference for orange fruits, and 0\% preference for green fruits. This behavior was not seen during pre-training. The final reward of the expert in this task is 40. 

\textbf{Highway-Env \& Roundabout-Env} \citet{highwayenv} features a collection of autonomous driving and tactical decision-making environments. We chose to model driving behavior as it allows us to determine the ability of IRL algorithms to learn the hidden intentions of a driver.
The ego-vehicle is driving on a road populated with other vehicles positioned at random. All vehicles can switch lanes and change their speed. We have modified the agent's reward objective to maintain a target speed, target distance from the front vehicle, and target lane, while avoiding collisions with neighbouring vehicles. As there are many continuous parameters that determine the reward function for the agent, it is not possible to sample all combinations of behaviors within the training tasks. Thus, it is straightforward to create a novel test task for the demonstrator by choosing a different combination of these behaviors.


\begin{figure}[t]
  \begin{center}
  \centerline{\includegraphics[width=.8\linewidth]{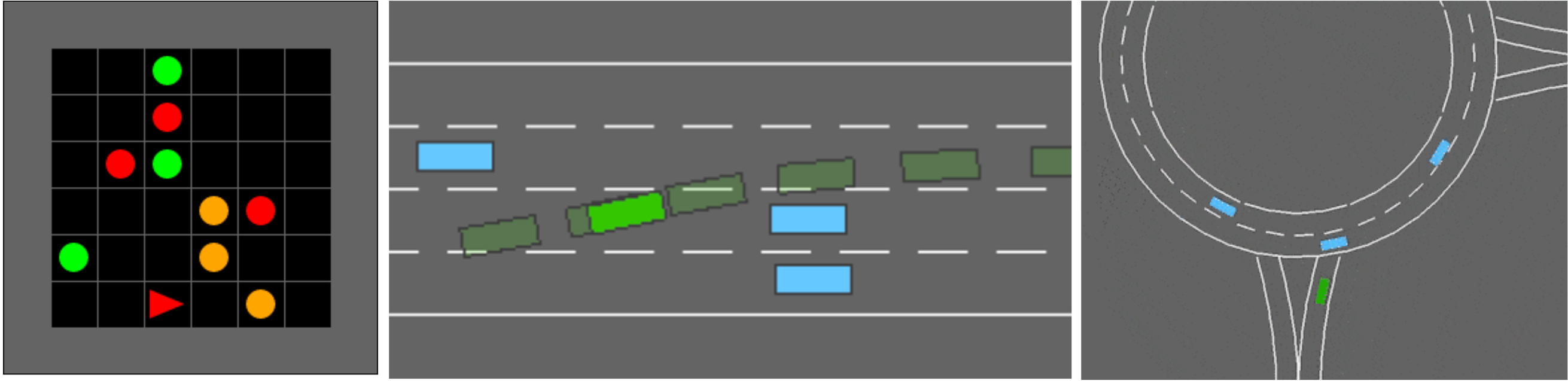}}
  \caption{We evaluate on the Fruit-Picking, Highway, and Roundabout domains (shown left to right).} 
  \label{fig:domain}
  \end{center}
\end{figure}

\subsection{Baselines and Comparisons}
We compare \acronym{} to three baselines. For all baselines, we use the same hyperparameters as \acronym{} when applicable, and maintain default values from source code otherwise. 
\textbf{IQ-learn} ~\citep{garg2021iqlearn} is a state-of-the-art, dynamics-aware, imitation learning (IL) method that is able to perform with very sparse training data, scaling to complex image-based environments. As it is able to implicitly learn rewards, this method can also be used in IRL. We use the authors' original implementation from: \textcolor{blue}{\url{https://github.com/Div99/IQ-Learn}}. 
\textbf{IQL (Inverse Q-learning)}~\citep{kalweit2020deep} is a state-of-the-art inverse RL method, which uses inverse-action value iteration to recover the underlying reward of an external agent, providing a strong comparison representative of recent inverse RL methods. 
\textbf{Multi-task IRL pre-training} is included to enable a fair comparison to methods which leverage multi-task pre-training. To our knowledge there is no prior method that uses RL pre-training to acquire a starting point for IRL. Closest in spirit is prior work on meta-IRL~\citep{yu2019metainverse, xu2019learning, gleave2018multitask},
which pre-train on other \emph{IRL} tasks, rather than pretraining on a set of standard RL problems, as in our method. 
Since these works generally assume a tabular or small discrete-state MDP, they are not directly applicable to our setting.
We thus create a multi-task IRL pre-training baseline by applying the same network architecture and BC and ITD losses as our own method (essentially performing our IRL phase (Algorithm \ref{alg:infer}) during pre-training as well). We note that in general, IRL pre-training has the significant disadvantage that it requires many more expert demonstrations during the pre-training phase, whereas our method does not. 




\section{Results}
We demonstrate \acronym's ability to leverage prior experience to help infer preferences quickly and efficiently on a diverse suite of domains. The code is available at 
\textcolor{blue}{\url{https://github.com/abdulhaim/basis-irl}} and the videos showing performance are available at \textcolor{blue}{\url{https://sites.google.com/view/basis-irl}}.

\begin{figure}
\centering
\begin{subfigure}{.25\linewidth}
  \centering
  \includegraphics[width=\linewidth]{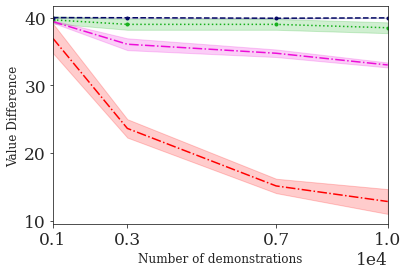}
  \caption{Fruit-Picking}
    \label{fig:fruitpick_value_comparison}
\end{subfigure}%
\begin{subfigure}{.25\linewidth}
  \centering
  \includegraphics[width=\linewidth]{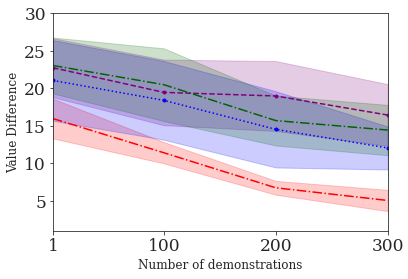}
  \caption{Highway}
  \label{fig:highway_value_comparison}
\end{subfigure}
\begin{subfigure}{.25\linewidth}
  \centering
  \includegraphics[width=\linewidth]{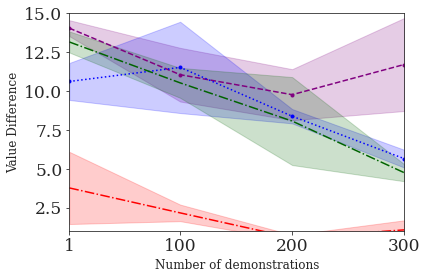}
  \caption{Roundabout}
  \label{fig:roundabout_value_comparison}
\end{subfigure}
\begin{subfigure}{.20\linewidth}
  \centering
  \includegraphics[width=\linewidth]{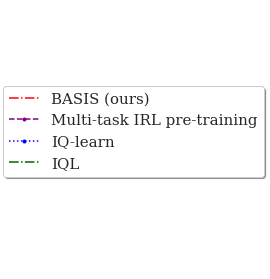}
\end{subfigure}
\begin{subfigure}{.25\linewidth}
  \centering
  \includegraphics[width=\linewidth]{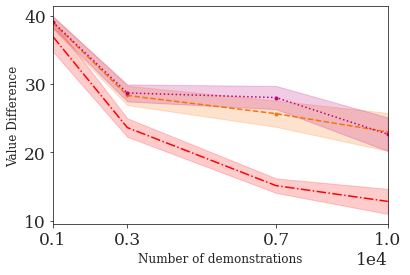}
  \caption{Fruit-Picking}
    \label{fig:fruitgrid_value_ablation}
\end{subfigure}%
\begin{subfigure}{.25\linewidth}
  \centering
  \includegraphics[width=\linewidth]{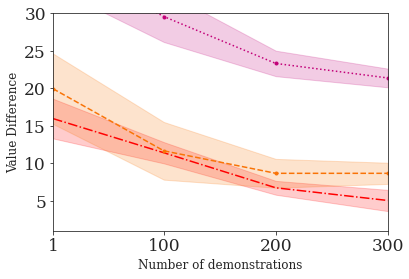}
  \caption{Highway}
  \label{fig:highway_value_ablation}
\end{subfigure}
\begin{subfigure}{.25\linewidth}
  \centering
  \includegraphics[width=\linewidth]{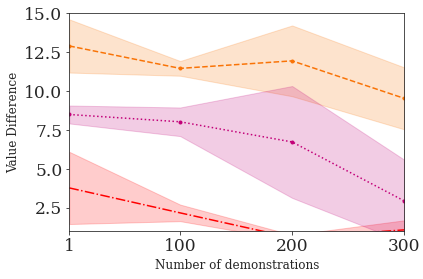}
  \caption{Roundabout}
    \label{fig:roundabout_value_ablation}
\end{subfigure}
\begin{subfigure}{.20\linewidth}
  \centering
  \includegraphics[width=\linewidth]{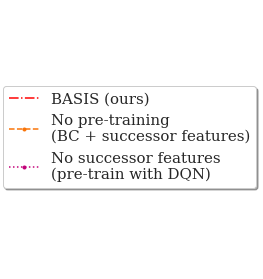}
\end{subfigure}
\caption{Value difference, i.e., the difference in the return obtained by each algorithm and the expert policy. Across both the Fruit-Picking (a), Highway (b), and Roundabout (c) domains, \acronym{} shows the lowest value difference indicating its behavior is closest to that of the optimal policy. It is able to surpass the performance of all baselines with less than one third of the data. The error bars show the standard deviation of 10 seeds.}
\label{fig:value_difference}
\end{figure}

\textbf{Question 1: Acquiring demonstrated behavior.}
Figure \ref{fig:value_difference} shows the value difference when evaluating an agent after learning from a series of demonstrations for all three environments. In Fruit-Picking, we observe that \acronym{} is better able to optimize the demonstrated policy than IQ-learn, IQL, and multi-task IRL pre-training, achieving the lowest value difference of 15 at 10000 demonstrations, and surpassing the final performance of the baselines with less than 1/3 of the demonstrations. 
We see similar trends in Figure \ref{fig:highway_value_comparison},
showing the value difference when inferring previously unseen driving preferences of desired driving distance, target speed, and lane preference in the Highway and Roundabout domains. The value difference for \acronym{} is seen to be significantly lower than all three baselines across all numbers of demonstrations, and it is once again able to surpass the best baseline with only 1/3 of the data requirements. 

\begin{figure}
\centering
\begin{subfigure}{.25\linewidth}
  \includegraphics[width=\linewidth]{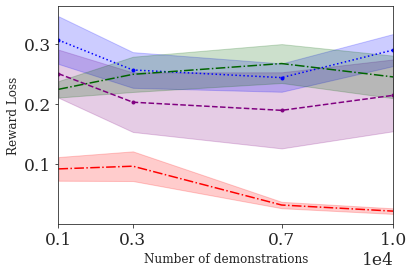}
  \caption{Fruit-Picking}
  \label{fig:fruitgrid_reward_comparisons}
\end{subfigure}%
\begin{subfigure}{.25\linewidth}
  \includegraphics[width=\linewidth]{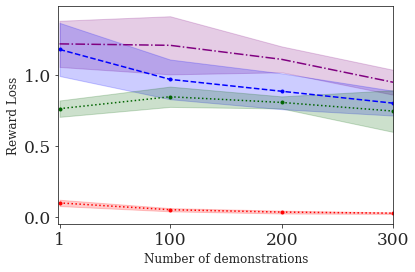}
  \caption{Highway}
  \label{fig:highway_reward_comparisons}
\end{subfigure}
\begin{subfigure}{.25\linewidth}
  \includegraphics[width=\linewidth]{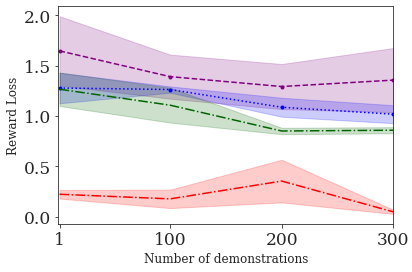}
  \caption{Roundabout}
  \label{fig:roundabout_reward_comparisons}
\end{subfigure}
\begin{subfigure}{.20\linewidth}
  \includegraphics[width=\linewidth]{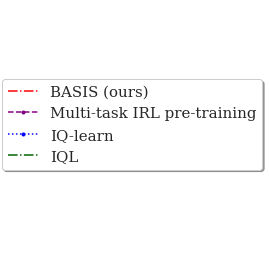}
\end{subfigure}
\begin{subfigure}{.25\linewidth}
  \includegraphics[width=\linewidth]{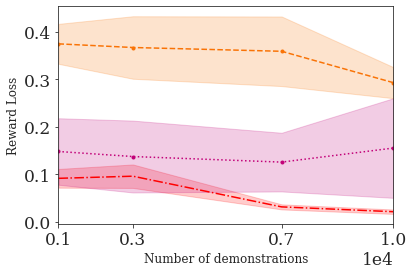}
  \caption{Fruit-Picking}
  \label{fig:fruitgrid_reward_ablation}
\end{subfigure}
\begin{subfigure}{.25\linewidth}
  \includegraphics[width=\linewidth]{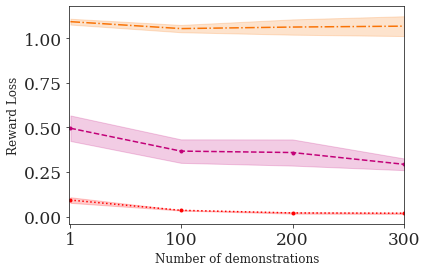}
  \caption{Highway}
  \label{fig:highway_reward_ablation}
\end{subfigure}
\begin{subfigure}{.25\linewidth}
  \includegraphics[width=\linewidth]{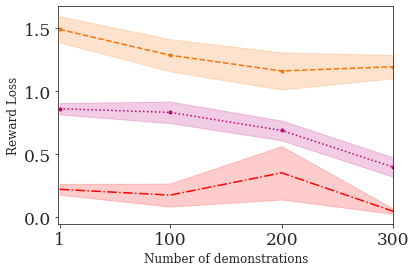}
  \caption{Roundabout}
  \label{fig:roundabout_reward_ablation}
\end{subfigure}
\begin{subfigure}{.20\linewidth}
  \includegraphics[width=\linewidth]{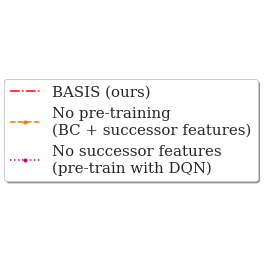}
\end{subfigure}
\caption{Reward loss: error in predicting the expert's true reward. \acronym{} is able to converge to a lower reward MSE loss compared to baselines in both the Fruit-Picking (a), Highway (b), and Roundabout (c) domains, which shows that it is most accurate in predicting the expert's reward. The error bars show the standard deviation of 10 seeds.}
\label{fig:reward_loss}
\vspace{-0.1cm}
\end{figure}

In the FruitGrid environment, we hypothesize that IQL as well as IQ-learn are unable to achieve a low value difference as shown in \cref{fig:fruitgrid_reward_comparisons},
due to the number of trajectories provided as well as a lack of ability to explore a range of rewards to discern the preferences of the agent.
We find that multi-task IRL shows a higher value difference,
as it was unable to reach maximum reward after an equivalent number of environment steps to multi-task RL in the pre-training phase. In addition, we note that conducting multi-task IRL pre-training requires collecting many expert trajectories, which may be prohibitively expensive, especially if these demonstrations are collected from humans. Instead, our method allows collecting experience inexpensively in simulation, making it much more sample efficient in terms of human demonstrations. 

This experiment allows us to determine whether \acronym{} fulfills the first requirement of being an IRL algorithm as well as an IL algorithm: being able to reproduce the behavior demonstrated by the expert. Further, because the performance of \acronym{} after only a few demonstrations surpasses the performance of both baselines after three times the number of demonstrations, this provides evidence that building a strong prior over the space of reasonable goals helps \acronym{} infer the expert's behavior rapidly and efficiently.


\textbf{Question 2: Predicting rewards.}
Figure \ref{fig:reward_loss} shows the reward loss (mean squared error in predicting the ground truth reward) obtained by each of the methods. Across all three environments, \acronym{} achieves the lowest error vs. any of the baseline techniques, often reaching the best performance in a fraction of the examples. 
In Fruit-Picking (\cref{fig:fruitgrid_reward_comparisons}), \acronym{} achieves the best performance after only 100 demonstrations, surpassing the performance of other techniques after 1000 demonstrations. Similarly, 
as shown in \cref{fig:highway_reward_comparisons}, we observe \acronym~to converge to a lower reward MSE loss than any of the other techniques after only one demonstration. This rapid inference of the expert's reward suggests that the basis acquired during pre-training was sufficient to explain the expert's behavior, and the algorithm was able to adapt rapidly to the expert's task by simply updating $w_e$ (as explained in Section \ref{sec:phaseii}). This is consistent with prior work using successor features for multi-agent learning \citep{filos2021psiphilearning}, which also showed 1-shot adaptation to a novel test task. 

\begin{figure}
\centering
\begin{subfigure}{.47\linewidth}
  \includegraphics[width=.95\linewidth]{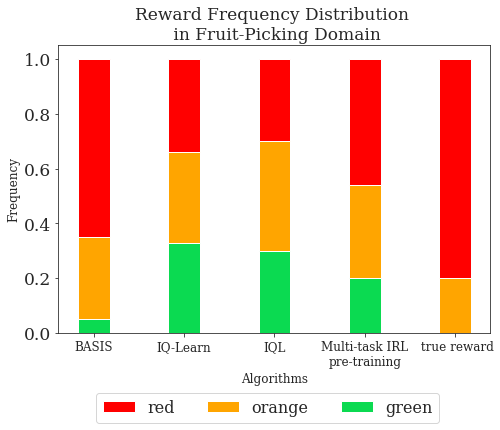}
  \caption{We test on a Fruit-Picking task where the agent must infer the demonstrator's
    preference to consume 80\% red fruits and 20\% orange fruits. \acronym{} is able to achieve this distribution more accurately than baseline techniques. We show the mean of our results over 10 seeds.} 
  \label{fig:fruit_distribution}
\end{subfigure}\hfill%
\begin{subfigure}{.47\linewidth}
\centering
  \includegraphics[width=.7\linewidth]{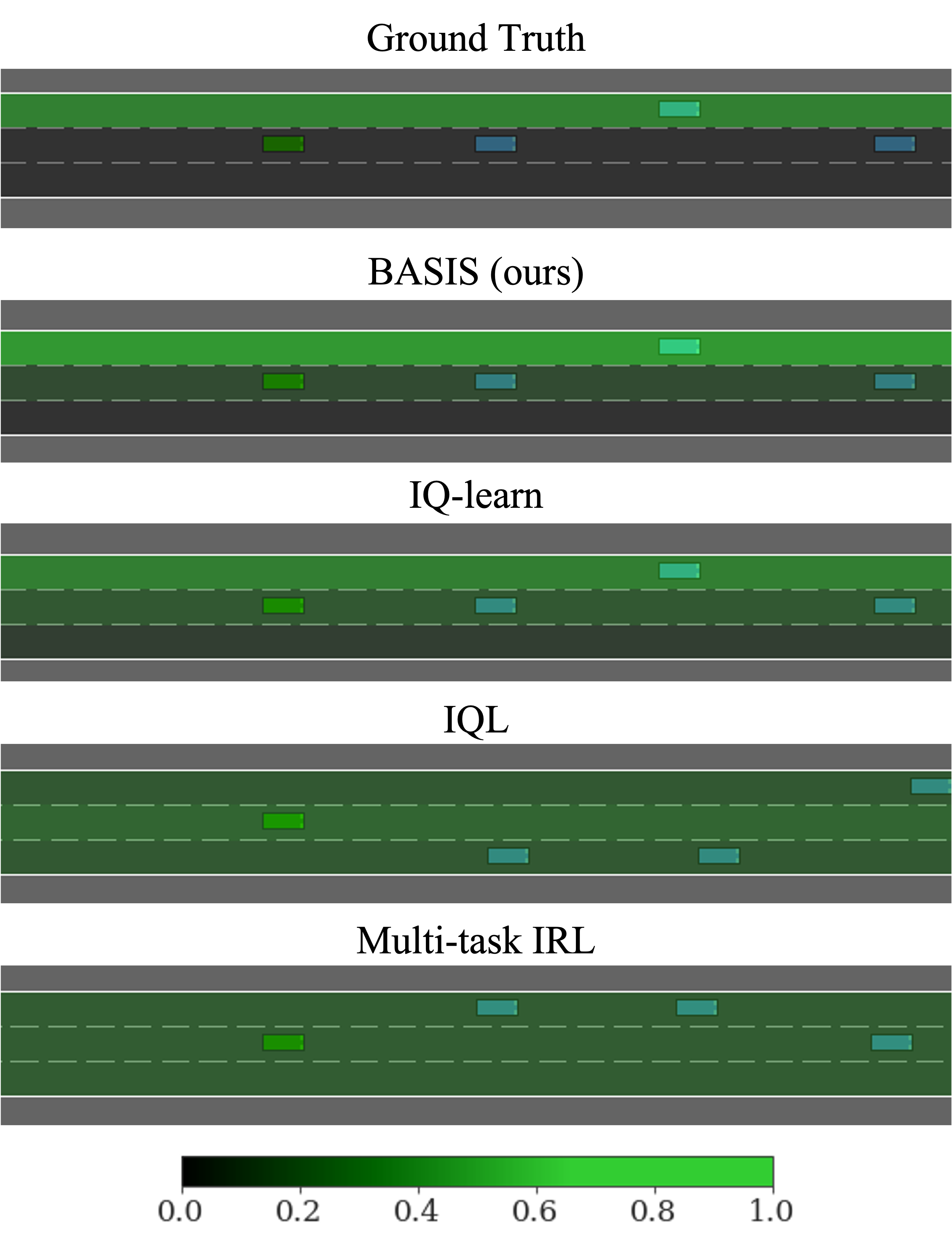}
  \caption{We show a state occupancy map for the degree of left lane preference in the Highway Domain. Our algorithm is best able to stick to the demonstrator agent's (ground truth) inferred preferences compared to baselines.}
  \label{fig:highway_state_occupancy}
\end{subfigure}
\caption{The above figures compare the distribution of behaviors learned by \acronym{} and each of the baseline techniques with the ground-truth behavior of the expert demonstrator.}
\label{fig:reward_studies}
\vspace{-0.1cm}
\end{figure}

\textbf{Question 3: Matching behavior statistics.}
To assess how well the different methods match the demonstrated behavior, we visually compare the distribution of behaviors for each technique to the expert's distribution. For the demonstrated task of collecting 80\% red fruit and 20\% orange fruit in the Fruit-Picking domain, we measure the distribution of fruits collected.
As shown in \cref{fig:fruit_distribution}, IQL gathers all fruits in roughly equal proportion, showing it has not learned to discern which fruits the expert prefers. This could be due to a failure to scale to high-dimensional environments. Although IQ-learn and Multi-task IRL pre-training are better able to capture the correct distribution, \acronym{} shows a distribution closer to the ground truth than either of the baseline techniques. 
 In Figure \ref{fig:highway_state_occupancy}, we perform a similar experiment in the Highway domain, visualizing how well the learned agent adheres to the staying in the preferred left lane. \acronym{} shows a preference for the left lane 80\% of the time, compared to IQ-learn which matches the expert's preference 60\% of the time. Both IQL and Multi-task IRL pre-training show an almost uniform lane preference, with no clear preference for the left lane. 

\textbf{Question 4: How does multi-task pre-training \& successor features benefit IRL?} 
We now conduct ablation experiments to assess how multi-task RL pre-training and successor features help in learning an IRL task in all three domains. We compare \acronym{} (in red) to two ablations: 1) \textit{No successor features} (which pre-trains a model with DQN), and 2) \textit{No pre-training} (which uses successor features and the BC loss in Eq. \ref{eq:bc}) to do IRL). We observe value difference in \cref{fig:value_difference} and reward loss in \cref{fig:reward_loss} to be larger without successor features and without multi-task pre-training, demonstrating the benefit of both components of our approach. We note that for both Fruit-Picking and Highway, the \textit{No pre-training} baseline does best, actually outperforming or matching the performance of all three baseline techniques (IQ-Learn, IQL, and multi-task IRL pre-training). However, \textit{No pre-training} performs poorly in the more complex Roundabout environment. In Roundabout, \textit{No successor features} actually gives better performance than two of the baseline techniques (IQL and multi-task IRL pre-training), demonstrating a strong benefit of learning a prior with RL pre-training.  This empirically shows how multi-task RL pre-training and successor features accelerate learning an IRL task.

\section{Conclusion}
A major challenge in inverse RL is that the problem is underconstrained. With many different reward functions consistent with observed expert behaviour, it is difficult to infer a reward function for a new task.
\acronym{} presents an approach to this problem by building a strong basis of intentions by combining multi-task RL pre-trainin with successor features. \acronym{} leverages past experience to infer the intentions of a demonstrator agent on an unseen task. We evaluate our method on domains with high dimensional state spaces, and compare to state-of-the-art inverse RL and imitation learning baselines, as well as pre-training with multi-task IRL. Our results show that \acronym{} is able to achieve better performance than prior work in less than one third of the demonstrations. The limitation of this approach is that it requires the design of a set of tasks for RL pre-training that are relevant to the expert demonstrations. Nevertheless, this work shows the potential of building a generalizable basis of intentions for efficient IRL.


\bibliography{iclr2022_conference}
\bibliographystyle{iclr2022_conference}
\section{Appendix}
\subsection{Environment details}
\label{sec:appendix}

\textbf{Fruit-Picking} is built from a library of open-source grid-world domains in sparse reward settings with image observations \citep{gym_minigrid}. We designed this custom environment with different colored fruits spread across the grid that the agent must gather. The number of fruits and types of fruits are customizable, along with the reward received for gathering a specific type of fruit. To learn a basis for intentions using RL pre-training, the agent learns to pick three different colored fruit (red, orange, green) depending on the task ID that is provided in the observation as a one-hot encoded vector. The agent picks as many fruits as it can until the horizon of the episode. Each fruit is replaced in a random location after being picked by the agent such that there are always 3 fruits of each color present in the grid. In Phase II, the demonstrated task is different to the training tasks; namely, the agent shows a varying degree of preference to each of the fruits in the environment i.e. 80\% preference for red fruits, 20\% preference for orange fruits, and 0\% preference for green fruits. This behavior was not seen during pre-training. The final reward of the expert in this task is 40. 

\textbf{Highway-Env \& Roundabout-Env} \citet{highwayenv} features a collection of autonomous driving and tactical decision-making environments. We chose to model driving behavior as it allows us to determine the ability of IRL algorithms to learn the hidden intentions of a driver.
In the highway env, the ego-vehicle is driving on a three-lane highway populated with other vehicles positioned at random. All vehicles can switch lanes and change their speed. We have modified the agent's reward objective to maintain a target speed while avoiding collisions with neighbouring vehicles and keeping to a preferred lane. Maintaining a target distance away from the front vehicle is also rewarded. As there are many continuous-spaced parameters that determine the reward function for the agent, it is not possible to sample all combinations of behaviors within the training tasks. Thus, to create a novel test task for the demonstrator, it is straightforward to choose a different combination of these behaviors. Specifically, we test on a task of the driving agent maintaining a desired distance 10 from the vehicle in front of it while maintaining a speed of 28 m/s.
We perform similar modifications to the Roundabout where the agent must merge onto a roundabout while maintaining a specific speed and target distance away from other vehicles. We build off the implementation of the Highway domain here: \textcolor{blue}{\url{https://github.com/eleurent/highway-env}}.

\subsection{Network architecture details}
For the MiniGrid Domain, networks for $\phi$, $\psi$ and $w$ policy share three convolution layers, with a ReLU after each layer and a max-pooling operation after the first ReLU activation. $\psi$ and $\phi$ are represented by two separate linear layers with a Tanh activation function between them. Finally, $w$ is represented as a single parameter. The network architecture is as follows: 
\begin{itemize}
\item Shared feature extraction layer:
    \begin{itemize}
        \item Conv2d (3, 16) 2x2 filters, stride 1, padding 0 
        \item ReLU 
        \item MaxPool2d (2, 2)  
        \item Conv2d (16, 32) 2x2 filters, stride 1, padding 0  
        \item ReLU  
        \item Conv2d (32, 64) 2x2 filters, stride 1, padding 0 
        \item ReLU  
    \end{itemize}
\item $\phi$ network layer: 
    \begin{itemize}
        \item FC (256 + num\_tasks, 64) 
        \item Tanh
        \item FC (64, num\_cumulants + num\_actions) 
    \end{itemize}
\item $\psi$ network layer:
    \begin{itemize}
        \item FC (256 + num\_tasks, 64) 
        \item Tanh
        \item FC (64, num\_cumulants + num\_actions) 
    \end{itemize}
\item $w$ network layer:  
    \begin{itemize}
        \item Parameter (num\_tasks, num\_cumulants)  \\
    \end{itemize}
\end{itemize}

In this domain, num\_cumulants=64, num\_actions=4, and num\_tasks=3. Note that the network architecture stays the same across RL pre-training and IRL, however during IRL, the num\_tasks hyper-parameter is not provided, and a dummy value (a vector of 0's equal to number of tasks) is used.

We now present the architecture for the Highway Domain. Networks for $\phi$, $\psi$ and $w$ policy share a linear layer with a ReLU activation after. $\psi$ and $\phi$ are represented by two separate linear layers with a ReLU activation function between them. Finally, $w$ is represented as a single parameter. The network architecture is as follows: 
\begin{itemize}
\item Shared feature extraction layer:
    \begin{itemize}
        \item FC (5, 256) 
        \item ReLU 
    \end{itemize}
\item $\phi$ network layer: 
    \begin{itemize}
        \item FC (1280 + num\_tasks, 256) 
        \item ReLU
        \item FC (256, num\_cumulants + num\_actions) 
    \end{itemize}
\item $\psi$ network layer:
    \begin{itemize}
        \item FC (1280 + num\_tasks, 256) 
        \item ReLU
        \item FC (256, num\_cumulants + num\_actions) 
    \end{itemize}
\item $w$ network layer:  
    \begin{itemize}
        \item Parameter (num\_tasks, num\_cumulants)  \\
    \end{itemize}
\end{itemize}

In this domain, num\_cumulants=64, num\_actions=5, and num\_tasks=10. Similar to Fruit-picking, the network architecture stays the same between RL pre-training and IRL.

\subsection{Implementation Details}

\begin{figure}[ht]
  \begin{center}
  \centerline{\includegraphics[width=0.40\linewidth]{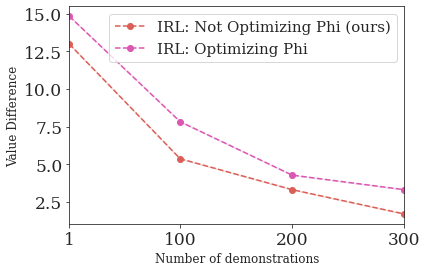}}
  \caption{Optimizing $\phi$ (IRL) causes no significant change.} 
  \label{fig:phi_changing}
  \end{center}
\end{figure}

\textbf{On global feature $\phi$}. We would like to clarify assumptions made in this paper that will address the reviewer's question on why $\phi$ transfers to the task during IRL (inferring intentions). As $\phi$ is indeed critical for the method's ability to recover effective rewards on downstream tasks, it is learned from the pre-training tasks which are within the same distribution (noted in Section 4). Our training procedure ensures that the $\phi$ features are sufficient to represent the pre-training tasks (optimized directly in the objective for $\psi$ and $\phi$). If these features then generalize to other in-distribution tasks in pre-training, they should also be sufficient for downstream tasks during IRL from the same distribution. Hence, we reuse $\phi$ during IRL and do not optimize it further. We have run an additional empirical analysis as suggested in Figure \ref{fig:phi_changing}, where we initialize $\phi$ from RL pre-training (learning a basis) and allow it to be optimized via ITD loss during IRL as well.There is marginal change in the value difference when optimizing and not optimizing $\phi$ (ours) during IRL in the Highway domain, confirming our intuition. \\ 

\textbf{On global feature $\phi$}. We would like to clarify assumptions made in this paper that will address the reviewer's question on why $\phi$ transfers to the task during IRL (inferring intentions). As $\phi$ is indeed critical for the method's ability to recover effective rewards on downstream tasks, it is learned from the pre-training tasks which are within the same distribution (noted in Section 4). Our training procedure ensures that the $\phi$ features are sufficient to represent the pre-training tasks (optimized directly in the objective for $\psi$ and $\phi$). If these features then generalize to other in-distribution tasks in pre-training, they should also be sufficient for downstream tasks during IRL from the same distribution. Hence, we reuse $\phi$ and do not optimize it further. We have run an additional empirical analysis as suggested in Figure \ref{fig:phi_changing}, where we initialize $\phi$ from RL pre-training (learning a basis) and allow it to be optimized via ITD loss during IRL as well.There is marginal change in the value difference when optimizing and not optimizing $\phi$ (ours) during IRL in the Highway domain, confirming our intuition. \\ 

\subsection{MaxEntropy Derivation}
Expanding on our explanation in Section 4.2, if $r = \phi(s,a)^Tw$ is the representation for the reward, and $Q(a,s) = \psi(a,s)w$, then the MaxEnt IRL problem can be written as:
\begin{equation}
\begin{split}
    \label{eq:derivation}
\max_{w,\phi} \E[\log \pi(a|s)] &= \max_{w,\phi}\E[\softmax(Q(s,a))] \\
 &= \max_{w,\phi} \E[\softmax(\psi(s,a)^T w)] \\
s.t. \quad \psi(s,a) = \phi(s,a) + &\gamma*\E_{a' ~ \softmax(\psi(s',a')^{\top} w)}[\psi(s',a')] 
\end{split}
\end{equation}

This leads to our method, which relaxes the constraint into a soft constraint.

\end{document}